\documentclass{article}

\usepackage[preprint,nonatbib]{neurips_2024}

\usepackage[utf8]{inputenc} % allow utf-8 input
\usepackage[T1]{fontenc}    % use 8-bit T1 fonts
\usepackage{hyperref}       % hyperlinks
\usepackage{url}            % simple URL typesetting
\usepackage{booktabs}       % professional-quality tables
\usepackage{amsfonts}       % blackboard math symbols
\usepackage{nicefrac}       % compact symbols for 1/2, etc.
\usepackage{microtype}      % microtypography
\usepackage{xcolor}         % colors
\usepackage{anyfontsize}    % any font size
\usepackage{tiaStyleFile}
\usepackage{tikz}           % Draw network
\usepackage{wrapfig}        % image in text
\usepackage{siunitx}

\title{Sparse $L^1$-Autoencoders for \\ Scientific Data Compression}

\author{%
  Matthias Chung\thanks{} \\
  Department of Mathematics\\
  Emory University\\
  Atlanta, GA 30322, USA \\
  \texttt{matthias.chung@emory.edu} \\
  \And
  Rick Archibald\\
  Computer Science and Mathematics Division\\
  Oak Ridge National Laboratory\\
  Oak Ridge, TN 37830, USA\\
  \texttt{archibaldrk@ornl.gov} \\
  \And
  Paul Atzberger\\
  Department of Mathematics\\
  University of California Santa Barbara \\
  Santa Barbara, CA 93106, USA\\
  \texttt{atzberg@gmail.com} \\
  \And
  Jack Michael Solomon  \\
  Department of Mathematics\\
  Emory University\\
  Atlanta, GA 30322 \\
  \texttt{jack.michael.solomon@emory.edu}
}

\begin{document}

\maketitle

\begin{abstract}
Scientific datasets present unique challenges for machine learning-driven compression methods, including more stringent requirements on accuracy and mitigation of potential invalidating artifacts.  Drawing on results from compressed sensing and rate-distortion theory, we introduce effective data compression methods by developing autoencoders using high dimensional latent spaces that are $L^1$-regularized to obtain sparse low dimensional representations.  We show how these information-rich latent spaces can be used to mitigate blurring and other artifacts to obtain highly effective data compression methods for scientific data. We demonstrate our methods for short angle scattering (SAS) datasets showing they can achieve compression ratios around two orders of magnitude and in some cases better. Our compression methods show promise for use in addressing current bottlenecks in transmission, storage, and analysis in high-performance distributed computing environments.  This is central to processing the large volume of SAS data being generated at shared experimental facilities around the world to support scientific investigations.  Our approaches provide general ways for obtaining specialized compression methods for targeted scientific datasets.
\end{abstract}

\section{Introduction \& Background}\label{sec:introduction}

Autoencoders are one of the most prominent and highly successful types of neural networks \cite{bourlard1988auto, kramer1992autoassociative,goodfellow2016deep}, they are attractive by their conceptual simplicity (learning parameterized encoder $e$ and decoder $d$ such that $x\approx d(e(x))$), their strong connection to well-established mathematical concepts \cite{plaut2018principal}, and versatile when utilized as generative models \cite{kingma2013auto}. Autoencoders have wide applicability in unsupervised learning environments ranging from denoising \cite{vincent2008extracting}, anomaly detection \cite{sakurada2014anomaly}, image and audio compression \cite{theis2022lossy,kumar2024high}, to generic recommender systems \cite{zhang2020survey}.

Most commonly, with their inherent structure of low dimensional latent spaces, autoencoders are a natural and data-driven machine learning technique for model reduction and data compression \cite{cheng2018deep,liu2021exploring}. Autoencoders' nonlinear characteristics make them a valuable complement to established techniques like principal component analysis and singular value decomposition (\cite{wang2016auto}), Fourier analysis (\cite{lappas2021fourier}), reduced order models (\cite{wu2021reduced}), and dictionary learning approaches (\cite{tariyal2016deep}).

Mathematically, an autoencoder is a nonlinear parameterized mapping $a:\calX \to \calX$ with a functional composition $a = d \circ e$, into \emph{encoder} $e:\calX \times \Theta_e \to \calZ$ and \emph{decoder} $d:\calZ \times \Theta_d \to \calX$ with trainable parameters $\theta_e \in \Theta_e \subset \bbR^{n_e}$ and $\theta_d \in \Theta_d\subset \bbR^{n_d}$. The feature space $\calZ\subset \bbR^\ell$ is referred to as \emph{latent space} while $\calX \subset \bbR^n$ is the \emph{data space}.

While terminology varies, autoencoders are classified based on their model configuration. Standard autoencoders with small dimensional latent spaces $\ell< n$ are referred to as \emph{undercomplete} and are widely utilized, while autoencoders with large dimensional latent spaces $\ell> n$ are referred to as \emph{overcomplete} and are less common. A \emph{shallow} autoencoder is characterized by having only one, while a \emph{deep} autoencoder has more than one hidden layer. Autoencoders typically maintain network symmetry, i.e., the structure of $e$ and $d$ are mirrored such that the decoder transformations mimic the encoder in reverse order. Its general structure makes autoencoders flexible, for instance, giving up the mapping back into the input space has led to encoder-decoder networks which are widely used as likelihood-free surrogate models for physical forward propagation \cite{zhu2019physics} and even inverse modeling~\cite{afkham2023goal}.

Despite its versatility, various drawbacks associated with autoencoders have been identified.  For instance, the undercomplete ``hour-glass'' autoencoder shape compresses input signals $x\in \calX$ into the low-dimensional latent space $z = e(x;\theta_e) \in \calZ$ and may lead to corrupted reconstructions, e.g., blurring artifacts in the reconstructed images \cite{meng2017relational}. This drawback has been highlighted in the generative process of autoencoders, i.e., variational autoencoders. Autoencoders with a large number of trainable network parameters, such as overcomplete autoencoder, tend to overfit, resulting in ``identity mappings'' countering one of the main purposes of autoencoders: removing unwanted artifacts from the input $x$ \cite{kabil2022undercomplete}.  A further drawback includes incorporating scientific and physical features into the network remains a major hurdle, where some initial research is making strides towards this goal \cite{lee2023nonlinear,cheng2024bi,bonneville2024comprehensive}.

\begin{figure}[b!]
    \centering
        \begin{tikzpicture}
	   \node[fill=matlab1!50, minimum width=0.5cm, minimum height=1.5cm] (x) at (0,0) {$x$};
	   \draw[fill=matlab3!50,draw=none] ([xshift=0.5cm]x.north east) -- ([xshift=2.5cm,yshift=1.5cm]x.east) -- ([xshift=2.5cm,yshift=-1.5cm]x.east) -- ([xshift=0.5cm]x.south east) -- cycle; 
	   \node at (1.75,0) {$\begin{matrix}\mbox{encoder} \\ e \end{matrix}$};
	   \node[fill=matlab5!50, minimum width=0.5cm, minimum height=3.0cm] (Zx) at (3.5cm,0) {$z$};
 	  \draw[fill=matlab2!50,draw=none] ([xshift=0.5cm]Zx.north east) -- ([xshift=2.5cm,yshift=-0.75cm]Zx.north east) -- ([xshift=2.5cm,yshift=0.75cm]Zx.south east) -- ([xshift=0.5cm]Zx.south east) -- cycle;
	   \node at (5.25,0) {$\begin{matrix}\mbox{decoder} \\ d \end{matrix}$};
	   \node[fill=matlab1!50, minimum width=0.5cm, minimum height=1.5cm] (B) at (7,0) {$x$};
    \end{tikzpicture}
    \caption{Overcomplete autoencoder architecture, where the latent space dimension $\ell$ is bigger than the input dimension $n$. Sparsity on the latent variable is imposed via $L^1$-type regularization.} \label{fig:sparseAutoencoder}
\end{figure}
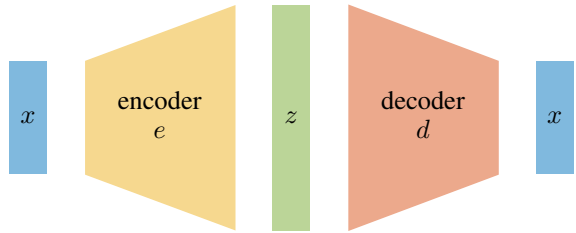

To mitigate the challenges outlined above, we break with one common assumption of autoencoders, that is, we consider utilizing an overcomplete autoencoder framework with sparsity promoting mappings of the latent variable, illustrated in \Cref{fig:sparseAutoencoder}. Overcomplete autoencoders are prone to severely overfit without taking mitigating measures. Hence, imposing sparsity onto the latent space variable is a regularizing measure and various strategies have been proposed. Despite being tremendously successful, we recognize that overcomplete autoencoders with sparsity-promoting features in the latent space are severely underutilized.

\emph{Sparse autoencoders} have first been introduced in the 2010s with pioneering work from various research groups including \cite{Ng2011SparseAutoencoder,Jiang2013SparseAutoencoder, makhzani2013k,majumdar2018autoencoder}. Here, various strategies have been entertained to promote sparsity of the latent variable. For instance, by selecting a fixed amount of $k$ nonzero elements of the latent variable $z$ with maximal reconstruction features \cite{makhzani2013k}. Another approach limits the number of active latent components by utilizing a binary Bernoulli random variable model realized through a Kullback-Leibner divergence penalty \cite{Ng2011SparseAutoencoder}. A third approach, which we will follow here, is to use the compressed sensing framework via $L^1$ regularization \cite{Jiang2013SparseAutoencoder}. Recent years have brought advances, various extensions of sparse autoencoders have been developed, and scientific applications considered \cite{kabil2022undercomplete, nguyen2020deep, martino2023we, kabil2022undercomplete, li2022deep, louizos2017learning, scardapane2017group, bricken2023towards}, however, despite its successes sparse autoencoder have yet to find its way into mainstream applications. We like to point out that the term sparse autoencoders may not refer to sparsity induced onto the the latent variable, but sparsity imposed on the network parameters  $\theta$, e.g., see \cite{louizos2017learning,scardapane2017group}.

To the best of our knowledge, a core application on the compressive feature of sparse autoencoder has not yet been fully addressed. Hence our proposed method utilizes sparse latent space signals to efficiently store input signals. Furthermore, compared to prior work, we consider not only the sparsity of the latent variable $z$ through $L^1$ regularization but promote sparsity on a signal $f(z)$. The functional $f$ provides the possibility to promote structure within the latent variable $z$, e.g., through a total variation type (generalized lasso).

Sparse autoencoder shares limitations of the general class of autoencoder, that is, the ability to effectively generalize to novel data instances, particularly when the training dataset does not accurately reflect the characteristics of the testing dataset.

Our work is structured as follows, in \Cref{sec:method} we introduce our proposed sparsity promoting autoencoder for data compression tasks, discuss its numerical applications in \Cref{sec:numerics}, and provide concluding remarks and discuss future work in \Cref{sec:conclusion}

\section{Sparse Autoencoders for Scientific Data Compression}\label{sec:method}

Scientific datasets present distinct challenges for machine learning-driven compression methods given how they are used applications.  Scientific investigations often require more stringent requirements on individual sample reconstruction accuracy and mitigation of artifacts such as blurring.  To help address these challenges, we use large dimensional information-rich latent spaces that are reduced by using sparsity regularizations to obtain representations amenable to further compression.  Historically, embedding sparse signals into large dimensional vector spaces has had a major impact on signal processing starting in the 1990s with the
\emph{compressed sensing framework} \cite{candes2005decoding,candes2006stable,tibshirani1996regression,donoho2006most}. We further develop this strategy for autoencoders using latent space dimensions larger than the feature space of the data, e.g., $\ell>m$.  Without imposing additional restrictions, learning the encoder $e(\mdot; \theta_e)$ and decoder $d(\mdot; \theta_d)$ would be an ill-posed problem without advantages for later compression.   Instead, we leverage the large dimensional spaces to allow our encoders to further process information and obtain well-posed compressed signals for the data $x\in \calX$ in $\calZ$.  We enforce sparsity on features of the latent vector $z$. Let $\theta_e$ and $\theta_d$ be the trainable network parameter of the encoder and decoder, respectively, ideally, we may formulate the network training as
\begin{equation}\label{eq:sparsereconstruction}
    \min_{(\theta_e, \theta_d) \in \Theta_e \times \Theta_d } \quad \bbE \norm[0]{f[e(x;\theta_e)]}
\quad \mbox{subject to} \quad \norm[2]{d(e(x;\theta_e);\theta_d) -x} \leq
\delta,
\end{equation}
where $\bbE$ denotes the expectation over the data $x$, $\norm[2]{\mdot}$ the $L^2$-norm, $\norm[0]{e(x; \theta_e)}$ is defined as the cardinality of nonzero elements in $e(x;\theta_e))$, and $\delta>0$ represents a desired reconstruction quality. We further let $f:\calZ \to \calF$ be a predefined operator we refer to as the \emph{sparse structure selector}.

Solving~\eqref{eq:sparsereconstruction} is NP-hard and efficient approximation approaches need to be utilized, \cite{elad2010sparse}. Under the restricted isometry properties \cite{candes2005decoding}, we may reformulate using an $L^1$ convex relaxation of \eqref{eq:sparsereconstruction} which leads to the generalized lasso problem
\begin{equation}\label{eq:lasso_gen}
    \min_{\theta_e, \theta_d \in \Theta} \quad \bbE \norm[2]{d(e(x;\theta_e);\theta_d) - x}^2 + \lambda \norm[1]{f[e(x;\theta_e)]},
\end{equation}
for suitable sparsity enforcing $\lambda>0$, see \cite{candes2006stable,chung2023variable} for details. This can be viewed as a rate-distortion objective, where $\lambda \norm[1]{\,\cdot\,}$ serves as a measure of the compression rate %$\mathcal{R}$
and $\norm[2]{\,\cdot\,}$ for the reconstruction distortion %$\mathcal{D}$
~\cite{Cover2006}.

What are the benefits of introducing an operator $f$ and not directly inducing sparsity on the latent variable $z$ with a standard $L^1$ regularization on the latent variable, e.g., $\norm[1]{e(x; \theta_e)}$?  With sparsity enforced only on each component of $z$ individually, the latent variable carries only minimal interpretability or structure \cite{martino2023we}. Now, a mapping $f$ may enforce structure to the latent space variable $z$. The function $f$ may remedy this disadvantage and add additional geometric interpretability. For simplicity,  we use a common total variation regularization approach $f[\,\cdot\,] = \nabla(\,\cdot\,) $, where $\nabla$ is the gradient operation.  In practice, for finite dimensional spaces $\calZ$, we approximate this operation by the finite difference operator $(f[z])_{i} = (z_{i+1} - z_{i})/h$, e.g., with $h = 1$. This is also used to help in clustering of information in the latent space.  %In our experiments we utilized a total variation operation in the mapping into the space.

Only because our autoencoder has a large dimensional latent space with a large amount of network parameter our approach does not just encode and decode the each training image individually within the network. But the numerical results in \ref{sec:numerics} clearly show that our approach generalizes well to testing data. % and are with high compression rates in the latent space.%  well encoded as a sparse vector in our latent space.

Further, note that the regularization parameter $\lambda$ balances potential over- and underfitting. Small $\lambda$ values may generate autoencoder with identity mappings for training data but may not generate any sparsity within the sparse structure selector. On the other hand large $\lambda$ may produce significant sparsity while missing to reconstruct the input signal $x$.  Cross validation techniques are readily available for calibrating the hyperparameter $\lambda$ but is subject to further investigations.

Given a representative set of (unsupervised) training samples $\{x_j\}_{j = 1}^m$ we minimize the empirical generalized lasso
\begin{equation}\label{eq:lasso}
    \min_{\theta_e, \theta_d \in \Theta} \quad \tfrac{1}{m} \sum_{j = 1}^m \norm[2]{d(e(x_j;\theta_e);\theta_d) - x_j}^2 + \lambda \norm[1]{f(e(x_j;\theta_e))},
\end{equation}

Our developed methods leverage results in compressed sensing showing promise for having a
\begin{wrapfigure}{r}{.6\textwidth}
    \centering
    \includegraphics[width=.6\textwidth]{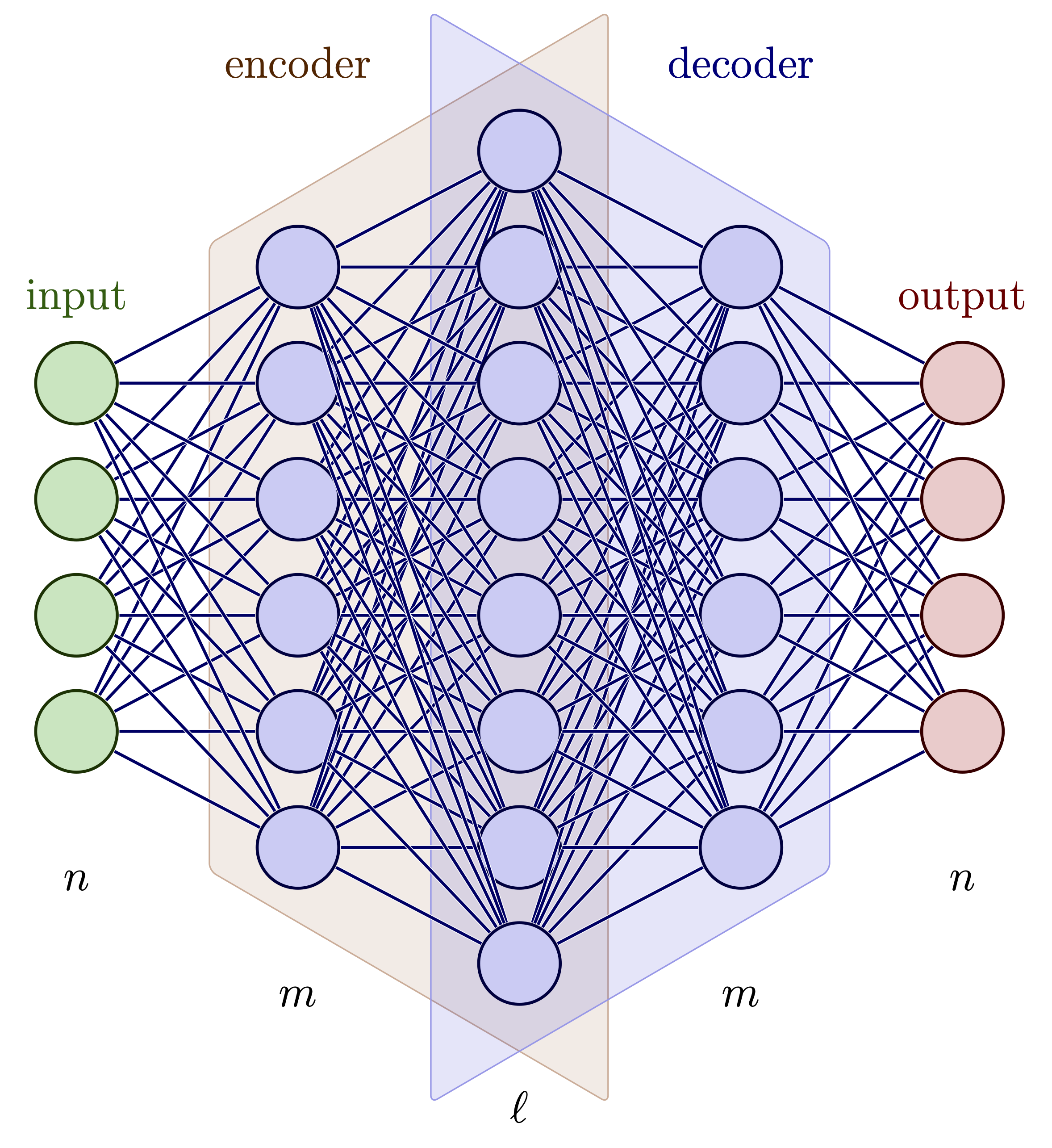}

    \caption{Autoencoder architecture for numerical examples. Network is a fully connected five layer symmetric neural network with hidden layer size $(m,\ell,m)$, $\relu$ activation function between each layer, and input/output size $n$.}\label{fig:tnfigure}
  \end{wrapfigure}
significant impact on scientific compression techniques. Under mild assumptions, compressed sensing has shown high compression rates, far below theoretical Nyquist rates \cite{candes2006robust,chen2010sub}. Benefiting from its advantages in a trainable deep neural networks provides significant compression rates while
maintaining high accuracy of the signal itself \Cref{sec:numerics}. Theoretically our methods have also connections to  dictionary learning frameworks. Sparse dictionary learning provides good reconstruction of a sparse selection of dictionary atoms \cite{tovsic2011dictionary,dumitrescu2018dictionary,kreutz2003dictionary,newman2023image}. Here, since linear, zeros in its dictionary representation does not carry any information. However, utilizing sparsifying autoencoders allows for nonlinear transformations and therefore enriches information carried by the latent variable. Consequently, even ``zero'' elements in the latent variable $z$ carry information of the underlying signal. Furthermore, the generic design of the neural network architecture provides an additional level of flexibility toward efficient encoding and decoding of the underlying signals $x$.
For (b), under mild assumptions, compressed sensing has shown high
compression rates, far below theoretical Nyquist rates
\cite{candes2006robust,chen2010sub}. Benefiting from its advantages in a
trainable deep neural networks may provide significant compression rates while
maintaining high accuracy of the signal itself.

We also develop methods to further compress our representations $z$ by combining our approaches with lossy quantization and lossless entropy encoding~\cite{gray1998quantization,Cover2006}.  We represent the information in $z$ as a list of the $m$ indices of nonzero entries $i_1,\ldots,i_m$ and the weights at these locations $w_k = z_{i_k}$.  We represent this by storing the distances between successive indices $\delta_k = i_{k+1} - i_k$ along with a termination symbol $\iota$ to obtain the sequence $\delta_1,\ldots, \delta_{m-1}, \iota$.  We expect in practice for most datasets that the probability distribution $\rho(\delta_i)$ over the differences $\delta_i$ will tend to skew toward the smaller values, such as having $\delta_i < \ell/2$ for most entries.  By modeling this distribution we can obtain gains in the compression using an entropy encoder.
To obtain a lossless entropy encoding for our model probability distribution $\rho(\delta)$, we develop a lossless arithmetic coding compression method $\mathcal{A}$ to obtain $c = \mathcal{A}(\delta;\rho(\cdot))$, ~\cite{witten1987arithmetic,langdon1984introduction,rissanen1979arithmetic}.
To compress the weights $\{w_k\}$ we use lossy quantization of the latent weights $\tilde{w} = \mathcal{Q}(w)$, such as using 16-bit
floating-points,~\cite{polino2018model,gray1998quantization,muller2018handbook}.  This provides for $z$ the compressed representation $(c,\tilde{w})$.  These methods provide further ways to compress the data in addition to the sparsity.

\section{Numerical Investigations}\label{sec:numerics}
\begin{figure}
    \centering
    \includegraphics[width = 1\textwidth]{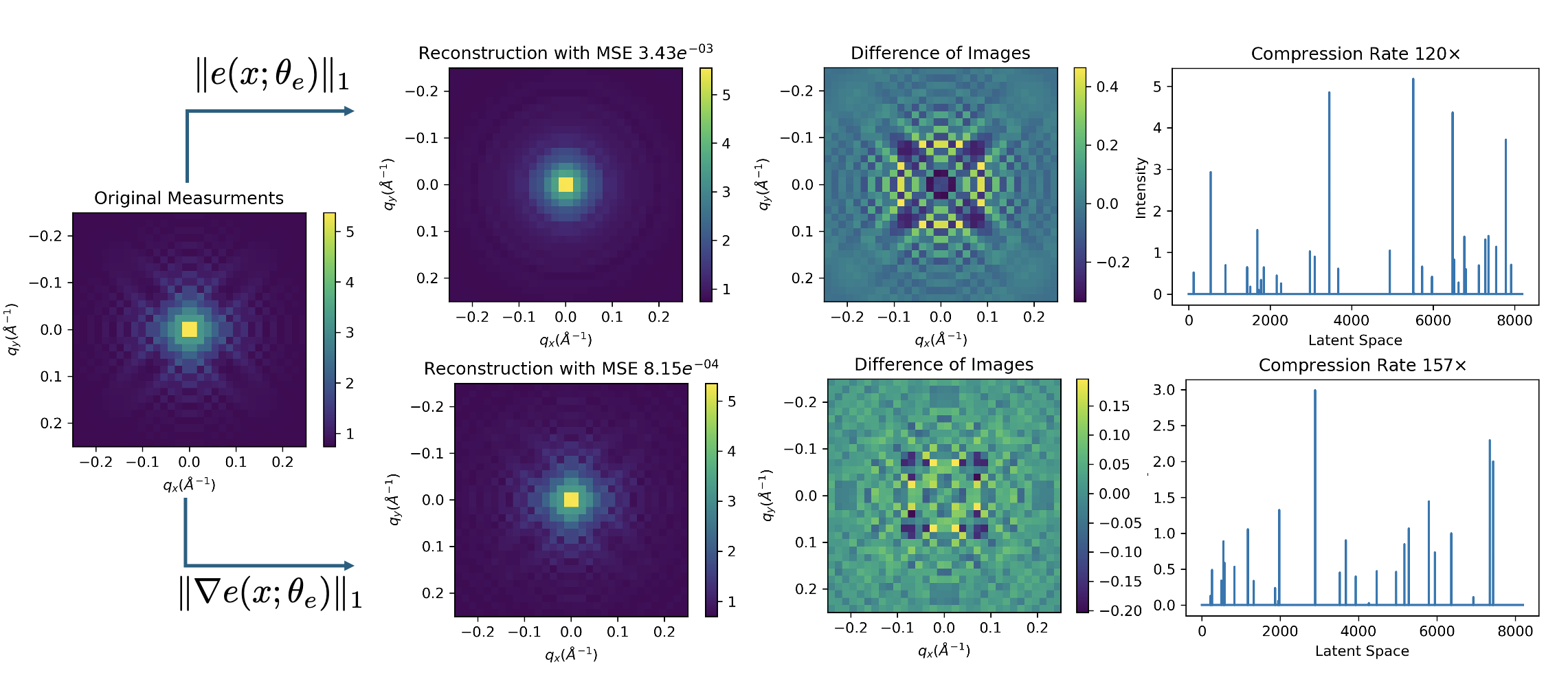}
    \caption{We show a representative testing input $x_j$ for the SAS application in the first column, while its reconstructions using sparse autoencoder networks $(1,2,10,z)$ on the top and $(1,2,10,\nabla z)$ on the bottom are presented in the second column. The reconstruction errors $\norm[2]{d(e(x_j; \theta_e); \theta_d) -  x}$ are $3.43\times 10^{-3}$ and $8.15\times 10^{-4}$, respectively. We show the latent space variable  $z_j = e(x_j; \theta_e)$ in the third column. The latent variable $z_j$ each contains 34 and 26  non-zero elements. With an original image size of $64\times 64$ the resulting compression ratio $\norm[0]{x_j}$ to $\norm[0]{e(x_j; \theta_e)}$ and  $\norm[0]{x_j}$ to $\norm[0]{\nabla e(x_j; \theta_e)}$ are $120:1$ and $157:1$.}
    % compression ratio
    \label{fig:sasresults}
\end{figure}
We illustrate the significant advantages our novel approach carries on simulated small angle scattering (SAS) data, a technique that is ubiquitous across the world's X-ray light and neutron facilities. We utilize the tool SASView \cite{sasview}, a community-based tool used at the experimental facilities to analyze and simulate SAS experiments. For all SAS experiments simulated, we set the number of sensors to be uniformly spaced with $n = 64\times64$. Measurement sensors for SAS experiments are some form of charge-coupled device (CCD), so uniform spacing is common. All networks in this section use the same autoencoder architecture depicted in \Cref{fig:tnfigure} and this architecture follows that depicted in \Cref{fig:sparseAutoencoder}. For a given input of size $n$ and loss defined in \Cref{eq:lasso}, we adopt the notation $\Big(\frac{m}{n},\frac{\ell}{n}, \lambda, f(z) \Big)$ to uniquely define all networks used in this paper. We further report that all networks were trained using $1,\!000$ epochs with a batch size of $512$ on Oak Ridge National Laboratory's (ORNL) Compute and Data Environment for Science (CADES) cluster \cite{cades}. In the initial phases of our investigation, we used convolution-type architectures which generally use convolution operators at the beginning and end of the network. We observed that the filtering aspect of the convolutional neural network dominated and hindered all information from reaching the latent space layer that connects the encoder and decoder.
\begin{figure}
    \centering
    \includegraphics[width = 1\textwidth]{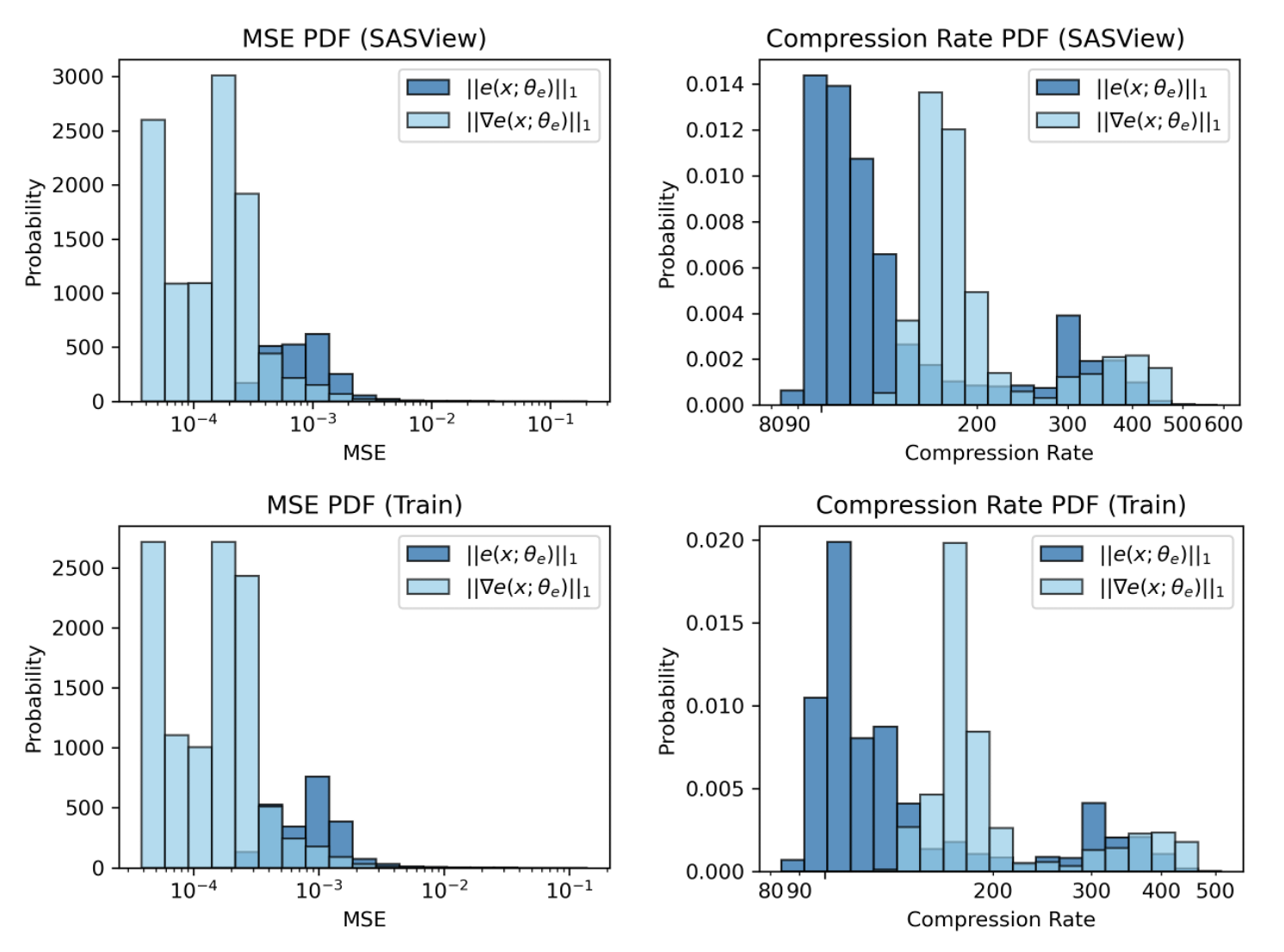}
    \caption{We show the error and compression rate partial distribution functions (PDFs) for the sparse autoencoder networks $(1,2,10,z)$ on the top and $(1,2,10,\nabla z)$ on the bottom. The top left plot displays the distributions of the entire training dataset, which consists of $50,\!000$ random SAS images generated by SASView. The bottom left plot displays the distributions in the prediction of $150,\!000$ independently random SAS images generated by SASView post-training. The mean training errors of both approaches are $2.48\times 10^{-3}$ and $8.00 \times 10^{-4}$, respectively. Correspondingly, the mean training compression rates are $153\times$ and $216\times$. Note that these values do only alter insignificantly for the testing set.}
    \label{fig:sasresults_dist}
\end{figure}

We demonstrate for realistic configurations of SASView \cite{HELLER2021100849}, we are able to highly compress all simulations from this package. We begin by randomly generating $50,\!000$ images using the aforementioned sensor configuration of $n = 64 \times 64$, which is characteristic of the range of SAS experimental data collected at scattering facilities. The representative result of this investigation is presented in \Cref{fig:sasresults}, which demonstrates high compression rates with high accuracy for the networks $(1,2,10,z)$ and $(1,2,10,\nabla z)$. The tested data averaged a relative reconstruction error of $7.75\cdot 10^{-2}$\%, and an averaged compression rate of $525\times$ with minimum rate $205\times$ and maximum rate $1365\times$. For comparison, we tested a similar fully connected encoder-decoder, with the same number of parameters, but had the typical hourglass framework without the sparsity promoting $L^1$ norm in~\eqref{eq:lasso}. With the same training and testing procedures, we were only able to obtain an average of $8\times$ compression rates with comparable accuracy. Using hyperparameter tuning, we found with the standard contracting encoder-decoder architecture the best size for the latent space was $\ell = 650$. This brings to light a significant difference with our sparse encoder-decoder, whereas the standard encoder-decoder requires hyperparameter tuning of the network architecture that results in training many different networks for set targets of accuracy and sparsity. Our sparse encoder-decoder is able to dynamically control the balance between sparsity and accuracy during training by adaptively changing the sparsity enforcing parameter $\lambda$ in~\eqref{eq:lasso} on a single network architecture.

We demonstrate that we have captured all realistic configurations of SASView \cite{HELLER2021100849}, using our train networks $(1,2,10,z)$ and $(1,2,10,\nabla z)$ in \Cref{fig:sasresults_dist}. Here we sample again a much denser set of measurements using $150,\!000$ images for testing. The takeaway from this analysis is that we can maintain the same level of compression and accuracy for both testing and training data. Additionally, it is demonstrated for the same number of network parameters, the sparsity promoting function $f(z) = \nabla z$ significantly improved compression rates and accuracy.  Again this is visually represented in \Cref{fig:sasresults} where this increased accuracy is able to maintain a more complex scattering pattern that can occur in SAS experiments.

\begin{figure}
    \centering
    \includegraphics[width = 1\textwidth]{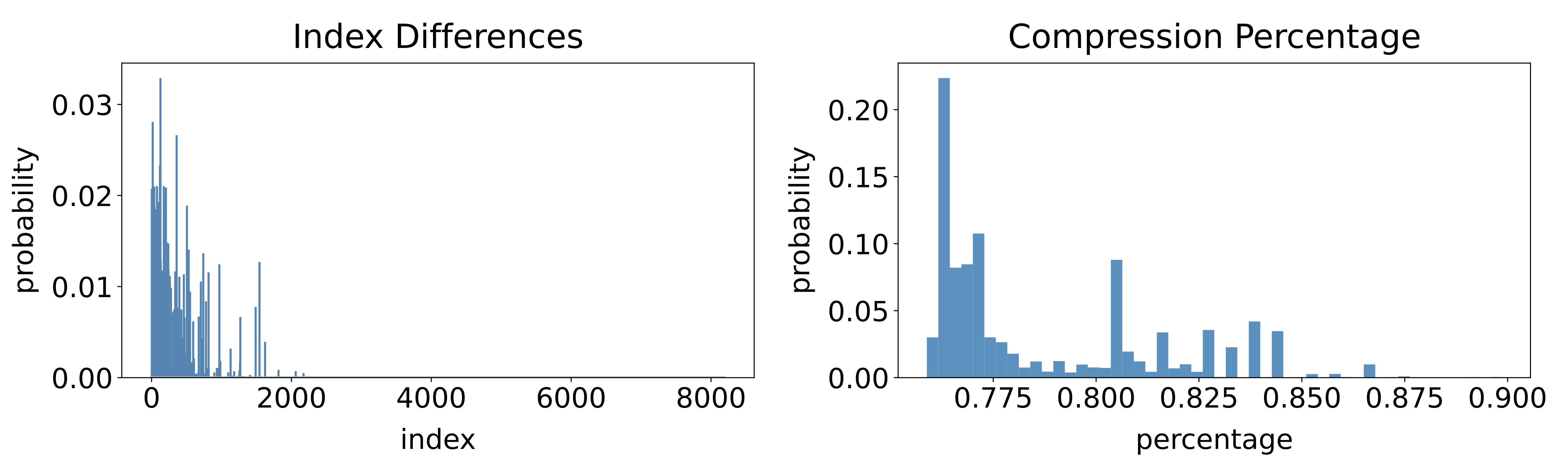}
    \caption{For further compression of $z$ with our arithmetic entropy encoding, we show the distribution of index differences $\rho(\delta_k)$ for our representation $z \rightarrow (\delta,w)$ (left).  For the SAS scattering data, we show the further compression reductions in percentage obtained for the index differences $\delta$ (right). }
    \label{fig:lossless_results}
\end{figure}

Our methods can also be combined with further lossy and lossless methods to obtain further compression.  As discussed above, we represent the information in $z$ as a list of differences in the $m$ indices of non-zero entries to obtain the sequence $\delta_1,\ldots, \delta_{m-1}, \iota$ and the weights $w_k$ at these indices to obtain $z \rightarrow (\delta,w)$.  Here, we use arithmetic coding $\mathcal{A}$ to develop for $\delta$ lossless compression methods $c = \mathcal{A}(\delta;\rho(\cdot))$~\cite{witten1987arithmetic,langdon1984introduction,rissanen1979arithmetic}. We also can quantize $\mathcal{Q}$ for lossy compression of $w$ as $\tilde{w} = \mathcal{Q}(w)$, such as using lower-precision floating-points~\cite{gray1998quantization,muller2018handbook}.
For $\delta$, we leverage that the probability distribution $\rho(\delta)$ will tend to skew to the left, for the SAS data see Figure~\ref{fig:lossless_results}.
As an initial model for this distribution, we use a Gaussian-like form $\rho(\delta) = q(\delta)/Z$, where $Z = \sum_\delta q(\delta)$ where $q(\delta)$ is normally distributed with density $
P(\delta;0,\sigma^2)$  where
$P(x;\mu,\sigma^2) = (2\pi \sigma^2)^{-1/2} \exp(-(x - \mu)^2/2\sigma^2)$.
To help ensure efficient encoding on future samples we used conservative parameters
$\sigma^2 = 10^3$ and $c_0 = 10^{-3}$.  We found the lossless compression methods provide on average a compressed representation $79\%$ of the uncompressed $\delta$.
Combining this with lossy quantization of the weights from $64$-bit floating-points to $16$-bit floating-points~\cite{polino2018model,gray1998quantization,muller2018handbook}, yields an overall compressed representation
$(c,\tilde{w})$ that is $52\%$ of the uncompressed case.  These methods provide an additional factor of around $2\times$ to the already favorable compression ratios achieved by the sparsity.

With respect to large scatter experiments across the world, there is starting to be a paradigm shift in how analysis of these measurements is being performed. Traditionally, analysis has been performed one experiment at a time on local clusters. However, as the datasets grow at these institutes, more complex analysis is needed at high-performance computing facilities that are not geographically collocated with experimental facilities. This presents a new challenge for scientific data reduction that requires guarantees of the reconstruction accuracy. Further, there is a goal of reducing analysis time, by doing this analysis in the compressed latent space. For this reason, we further investigate the properties of our sparse autoencoder using the well-studied MNIST (Modified National Institute of Standards and Technology database) database \cite{deng2012mnist}.
\begin{figure}[H]
    \centering
    \includegraphics[width = 1\textwidth]{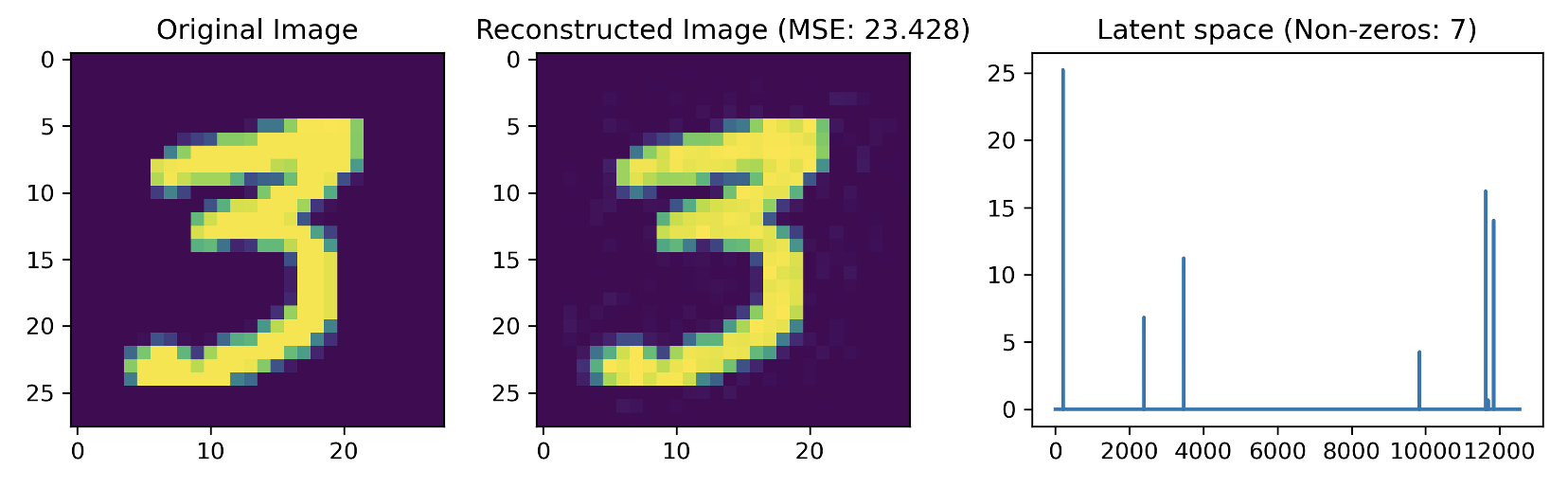}
    \caption{We show a representative testing input $x_j$ for the MNIST dataset in the first column at a compression with an MSE of 23.428 and a compression ratio of $120:1$.}
    \label{fig:mnist_rec}
\end{figure}
We first demonstrate the high level of compression possible using a sparse network $(8,16,5000,z)$ using the given $60,\!000$ testing and $10,\!000$ handwritten images of size $n=28\times 28$. The purpose of this shift of dataset attention is to determine if the latent space can be used to maintain classification accuracy and how latent space size affects this classification problem. Here, we are not concerned with the accuracy of any classification procedure per se, but rather the comparison of classification rates as a function of compression. For this reason, we consider the classification procedure of k-nearest neighbors (KNN) using the Cosine similarity measure. \Cref{{tab:knn}} gives a synopsis of the results.
\setlength\tabcolsep{1pt}
\begin{table}[h]
\fontsize{9pt}{9pt}\selectfont
\centering
\begin{tabular}{c|c|c|c|c|c|c|c}
\toprule
\multicolumn{1}{c}{\textbf{Architecture}} & \multicolumn{2}{c}{\textbf{$L^2$}} & \multicolumn{2}{c}{\textbf{$ L^1$}} & \multicolumn{2}{c}{\textbf{$L^0$}} & \\
\cmidrule(rl){2-3} \cmidrule(rl){4-5} \cmidrule(rl){6-7}
$\Big(\frac{m}{n},\frac{\ell}{n},\lambda ,z \Big)$ & \textbf{Test} & \textbf{Train} & \textbf{Test} & \textbf{Train} & \textbf{Test} & \textbf{Train} & \textbf{KNN} \\
\midrule

$(\nicefrac{1}{2},\nicefrac{1}{4},0,z)$ & \scriptsize\SI[round-mode=uncertainty,round-precision=1,tight-spacing=true]{194.36626631445537(16.31396850404305)}{} & \scriptsize\SI[round-mode=uncertainty,round-precision=1,tight-spacing=true]{181.44215851547605(16.628382796221725)}{} & \scriptsize\SI[round-mode=uncertainty,round-precision=1,tight-spacing=true]{182.9805309564647(13.553313978029767)}{} & \scriptsize\SI[round-mode=uncertainty,round-precision=1,tight-spacing=true]{182.65494334457034(13.398911311683085)}{} & \scriptsize\SI[round-mode=uncertainty,round-precision=1,tight-spacing=true]{156.92486453073943(1.1343895309028733)}{} & \scriptsize\SI[round-mode=uncertainty,round-precision=1,tight-spacing=true]{156.97369047619046(1.1730617597936817)}{} & \scriptsize\SI[round-mode=uncertainty,round-precision=1,tight-spacing=true]{0.954(0.002489424087253785)}{} \\
$(\nicefrac{3}{4},\nicefrac{1}{2},50,z)$ & \scriptsize\SI[round-mode=uncertainty,round-precision=1,tight-spacing=true]{159.9066767428395(12.55383797582193)}{} & \scriptsize\SI[round-mode=uncertainty,round-precision=1,tight-spacing=true]{127.23925809735344(12.162124821250986)}{} & \scriptsize\SI[round-mode=uncertainty,round-precision=1,tight-spacing=true]{1.2036540275994654(0.0006258307619614891)}{} & \scriptsize\SI[round-mode=uncertainty,round-precision=1,tight-spacing=true]{1.1958500205753815(0.004533400669260136)}{} & \scriptsize\SI[round-mode=uncertainty,round-precision=1,tight-spacing=true]{159.00597475565232(4.646536311658809)}{} & \scriptsize\SI[round-mode=uncertainty,round-precision=1,tight-spacing=true]{157.22559523809525(4.6543361270375305)}{} & \scriptsize\SI[round-mode=uncertainty,round-precision=1,tight-spacing=true]{0.9420918786667934(0.0013764485208486766)}{} \\
$(1\nicefrac{1}{2},2,200,z)$ & \scriptsize\SI[round-mode=uncertainty,round-precision=1,tight-spacing=true]{215.13073713446593(15.660555740368022)}{} & \scriptsize\SI[round-mode=uncertainty,round-precision=1,tight-spacing=true]{164.7981555286816(18.402420796266114)}{} & \scriptsize\SI[round-mode=uncertainty,round-precision=1,tight-spacing=true]{0.1342488110980817(0.0022775413026790464)}{} & \scriptsize\SI[round-mode=uncertainty,round-precision=1,tight-spacing=true]{0.13369166601608906(0.0018649860824158189)}{} & \scriptsize\SI[round-mode=uncertainty,round-precision=1,tight-spacing=true]{106.97793720843322(4.890704048597305)}{} & \scriptsize\SI[round-mode=uncertainty,round-precision=1,tight-spacing=true]{105.76283333333333(4.63459407161153)}{} & \scriptsize\SI[round-mode=uncertainty,round-precision=1,tight-spacing=true]{0.9441056243321696(0.0009501164742078755)}{} \\
$(2,4,800,z)$ & \scriptsize\SI[round-mode=uncertainty,round-precision=1,tight-spacing=true]{362.13353735185956(126.57596552816511)}{} & \scriptsize\SI[round-mode=uncertainty,round-precision=1,tight-spacing=true]{255.64132300840106(86.69546186403318)}{} & \scriptsize\SI[round-mode=uncertainty,round-precision=1,tight-spacing=true]{0.08781843645304226(0.0025339039311551227)}{} & \scriptsize\SI[round-mode=uncertainty,round-precision=1,tight-spacing=true]{0.08750441303379125(0.0024431551446350485)}{} & \scriptsize\SI[round-mode=uncertainty,round-precision=1,tight-spacing=true]{67.24809327447744(40.36141249928829)}{} & \scriptsize\SI[round-mode=uncertainty,round-precision=1,tight-spacing=true]{66.60633333333334(39.79556027164924)}{} & \scriptsize\SI[round-mode=uncertainty,round-precision=1,tight-spacing=true]{0.9510502892073746(0.005718327997944749)}{} \\
$(4,8,3200,z)$ & \scriptsize\SI[round-mode=uncertainty,round-precision=1,tight-spacing=true]{308.999645786012(126.57596552816511)}{} & \scriptsize\SI[round-mode=uncertainty,round-precision=1,tight-spacing=true]{246.52328803(80.37161822)}{} & \scriptsize\SI[round-mode=uncertainty,round-precision=1,tight-spacing=true]{0.0191362(0.00111659)}{} & \scriptsize\SI[round-mode=uncertainty,round-precision=1,tight-spacing=true]{0.0191362514(0.00125489)}{} & \scriptsize\SI[round-mode=uncertainty,round-precision=1,tight-spacing=true]{51.232333333(8.79537591)}{} & \scriptsize\SI[round-mode=uncertainty,round-precision=1,tight-spacing=true]{52.6295(11.31584216)}{} & \scriptsize\SI[round-mode=uncertainty,round-precision=1,tight-spacing=true]{0.9568(0.00499786)}{} \\
\bottomrule
\end{tabular}
\vspace{.1cm}
\caption{This table displays the average $L^2$ norm for the difference of the input image and output image, $\norm[2]{d(e(x_j;\theta_e);\theta_d) - x_j}$, with the $L^1$ and $L^0$ norm for the latent space, $\norm[1]{f(e(x_j;\theta_e))}$ and $\norm[0]{f(e(x_j;\theta_e))}$, for different autoencoder architectures. The statistical notation, for example \SI[round-mode=uncertainty,round-precision=1,tight-spacing=true]{156.92486453073943(1.1343895309028733)}{}, implies $157\pm1$. The uncertainty was determined by training five autoencoders on a random sample of 20\% on all of the MNIST dataset (combining the $60,\!000$ training and $10,\!000$ testing into $70,\!000$ samples that were randomly split into $14,\!000$ training and $56,\!000$ testing). To measure the informational content of the latent space we report the KNN accuracy by training on the $14,\!000$ set latent space representation and testing on the derived $56,\!000$ latent space. For reference the KNN accuracy in the image space is \SI[round-mode=uncertainty,round-precision=1,tight-spacing=true]{0.9596411180851406(0.008696022043471449)}{}.}
\label{tab:knn}
\end{table}

We train multiple different sparse autoencoders on a random sample of 20\% on all of the MNIST dataset. We know for KNN on the full MNIST training set that KNN accuracy in the image domain is 97\%.  Multiple training and test of the KNN accuracy in the image space with a 20\% - 80\% split on MNIST dataset produces a benchmark distribution of \SI[round-mode=uncertainty,round-precision=1,tight-spacing=true]{0.9596411180851406(0.008696022043471449)}{}. \Cref {tab:knn} shows the reconstruction error and latent space compression rate for latent spaces that range from $\frac{\ell}{n}\in[\frac{1}{4}-8]$. For reference the network $(\nicefrac{1}{2},\nicefrac{1}{4},0,z)$ has no sparsity enforcement in the latent space, but rather sparsity is enforced through the hourglass architecture. The sparsity parameter $\lambda$ for every other row was set to approximate the reconstruction accuracy achieved by $(\nicefrac{1}{2},\nicefrac{1}{4},0,z)$. The KNN result was trained on by the latent space each sparse autoencoder trained on and tested on the derived testing latent space produced by the same sparse autoencoder. The takeaway from this analysis is that the compression rate is increasing as a function of latent space size. Accuracy of reconstruction is maintained for each sparse autoencoder on data that the network never trained upon, and finally KNN prediction accuracy in the latent space increases with latent space size. The implications of this to analysis on scientific data reduction are positive in the sense that given a small random sample of a scientific dataset, sparse autoencoders with large latent spaces can maximize compression rates and this compressed representation provides equivalent information for analysis in the latent space in comparison to analysis in the data space.

\section{Conclusion and Future Work}\label{sec:conclusion}

Our introduced sparse autoencoder methods provide natural extensions of compressed sensing approaches for use in the lossy compression of scientific data.  Our introduced sparsity-promoting regularizations on the mappings of the latent variable were demonstrated to provide significant benefits for encoding scientific short-angle scattering data.  Our numerical investigations indicate that the use of information-rich large dimensional latent spaces provides significant advantages in preserving features of signals during compression.  Our methods provide ways for obtaining sparse representations by separating the structure of the encoded signals from the latent variable representations.  Our methods introduce robust learning strategies enabling significant compression ratios allowing for the accurate and efficient storage, transmission, and analysis of scientific datasets.  Our introduced methods also can be combined with other autoencoder strategies and lossy/lossless compression methods for handling diverse types of data in scientific applications.

\paragraph{Acknowledgement:} This work was partially supported by the National Science Foundation (NSF) under grant  DMS-2152661 (M. Chung).  Author P.J.A. would like to acknowledge support from NSF Grant DMS-2306101. Author R.A. would like to acknowledge support the Office of Advanced Scientific Computing Research and performed at the Oak Ridge National Laboratory, which is managed by UT-Battelle, LLC for the US Department of Energy under Contract No. DE-AC05-00OR22725. Research used resources of the Oak Ridge Leadership Computing Facility at Oak Ridge National Laboratory, which is supported by the Office of Science of the U.S. Department of Energy under Contract No. DE-AC05-00OR22725

\printbibliography

\end{document}